\documentclass[a4paper, 10pt, conference]{ieeeconf}      

\usepackage{times}
\usepackage{epsfig}
\usepackage{graphicx}
\usepackage{amsmath}
\usepackage{amssymb}
\usepackage{textcomp}
\usepackage{multirow}
\usepackage{adjustbox}

\usepackage{multirow}
\usepackage{colortbl}
\usepackage{hhline}
\usepackage{subcaption}
\usepackage{verbatim}

\usepackage{gensymb}
\usepackage{flushend}
\usepackage{xcolor}
\usepackage{dirtytalk}
\usepackage{multirow}
\usepackage{hyperref}

\IEEEoverridecommandlockouts                              

\title{\LARGE \bf
TiledSoilingNet: Tile-level Soiling Detection on Automotive Surround-view Cameras Using Coverage Metric
}

\author{
Arindam Das$^{1}$, 
Pavel K\v{r}\'{i}\v{z}ek$^{2}$, 
Ganesh Sistu$^{3}$, 
Fabian Bürger$^{4}$,\\
Sankaralingam Madasamy$^{1}$, 
Michal U\v{r}i\v{c}\'{a}\v{r}$^{2}$, 
Varun Ravi Kumar$^{5}$ and 
Senthil Yogamani$^{3}$\\ 
$^{1}$Detection Vision Systems, Valeo India, 
$^{2}$Valeo R\&D Prague, Czech Republic,\\ 
$^{3}$Valeo Visions Systems, Ireland, 
$^{4}$Valeo DAR Bobigny, France and 
$^{5}$Valeo DAR Kronach, Germany\\
{\tt \small \{firstname.lastname\}@valeo.com}
}

\begin{document}

\maketitle
\thispagestyle{empty}
\pagestyle{empty}

\begin{abstract}
Automotive cameras, particularly surround-view cameras, tend to get soiled by mud, water, snow, etc. For higher levels of autonomous driving, it is necessary to have a soiling detection algorithm which will trigger an automatic cleaning system. Localized detection of soiling in an image is necessary to control the cleaning system. It is also necessary to enable partial functionality in unsoiled areas while reducing confidence in soiled areas. Although this can be solved using a semantic segmentation task, we explore a more efficient solution targeting deployment in low power embedded system. We propose a novel method to regress the area of each soiling type within a tile directly. We refer to this as coverage. The proposed approach is better than learning the dominant class in a tile as multiple soiling types occur within a tile commonly. It also has the advantage of dealing with coarse polygon annotation, which will cause the segmentation task. The proposed soiling coverage decoder is an order of magnitude faster than an equivalent segmentation decoder. We also integrated it into an object detection and semantic segmentation multi-task model using an asynchronous back-propagation algorithm. A portion of the dataset used will be released publicly as part of our WoodScape dataset \cite{yogamani2019woodscape} to encourage further research.
\end{abstract}

\section{INTRODUCTION}
Autonomous driving systems have become mature over time, especially in offering various driving assistance features. It has been possible with the help of various cost-effective and reliable sensors. Especially cameras, provide a lot of useful information while being relatively cheap. There is extensive progress in various visual perception tasks such as semantic segmentation~\cite{briot2018analysis}, moving object detection \cite{yahiaoui2019fisheyemodnet}, depth estimation \cite{kumar2018monocular}, re-localisation \cite{tripathi2020trained} and fusion \cite{rashed2019fusemodnet, rashed2019motion}. Due to the maturity of these visual perception algorithms, there is increasing focus on difficult conditions due to adverse weather conditions such as snow, rain, fog, etc. 

\begin{figure*}[tb]
    \centering
    \includegraphics[width=0.89\textwidth]{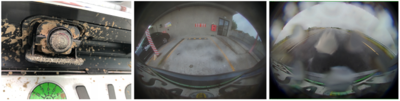}
    \caption{From left to right: a) soiled camera lens mounted to the car body; b) the image quality of the soiled camera from the previous image; c) an example image soiled by heavy rain.}
    \label{fig:saw_definition}
\end{figure*}

Some automotive cameras are attached outside of the car, and the camera lens can get contaminated by environmental sources such as mud, dust, dirt, grass, etc. It is a related but different problem compared to adverse weather handling. The main difference is a cleaning system can reverse that camera lens soiling. Soiling on lens causes severe degradation of the performance of visual perception algorithms, and it is necessary to have a reliable soiling detection algorithm that would trigger a cleaning system.  Failure in the detection of soiling can severely affect the automated driving features. Figure~\ref{fig:saw_definition} shows some samples of water drops and mud splashes on the camera lens.

Evaluation of vision algorithms in poor visibility conditions has received significant attention from the computer vision community. A competition called UG2 \cite{ug2} was organized by CVPR 2019, and a recent CVPR 2020 workshop focused on a vision for all seasons \cite{v4as}. Semantic segmentation for scenes with adverse weather situations was explored in \cite{sakaridis2018model, pfeuffer2019robust, alshammari2019multi}. Porav et al. \cite{porav2019i} created a setup to capture simultaneous scenes with and without rain effects by dripping water on one camera lens while keeping the second one clean. Image restoration approaches for Dehazing was proposed in \cite{Fattal-2008, Berman-2016, Ki-2018}. 

Relatively, there is limited work on the analysis of lens soiling scenarios, and we summarize existing work below. The problem of degraded visual perception due to soiled objects or adverse weather scenarios for automated driving was presented first in \cite{Uricar-2019a}, where data augmentation was performed using GAN (Generative Adversarial Network). SoilingNet \cite{uricar2019soilingnet} proposed a Multi-Task Learning (MTL) framework to train the decoder for soiling tasks, and it includes an overview of the camera cleaning system. The efficient soiling detection network called \emph{SoildNet} was explored in \cite{das2019soildnet} for embedded platform deployment.

\noindent The main contributions of this paper are as follows: 
\begin{enumerate}
    \item Design of a novel coverage based tile-level soiling detection method.
    \item Implementation of a highly efficient soiling coverage decoder, which is faster than segmentation decoders.
    \item Implementation of an asynchronous back-propagation algorithm for training of soiling and other semantic tasks.
\end{enumerate} 

The paper is organized as follows. Section \ref{sec:soiling} defines the coverage based soiling detection task formally and discusses its benefits over the alternatives. Section \ref{sec:proposed} discusses the dataset design, proposes a Convolutional Neural Network (CNN) architecture for soiling and the integration into a multi-task architecture. Section \ref{sec:results} presents the experimental analysis and an overview of the remaining practical challenges. Finally, Section \ref{sec:conc} summarizes and concludes the paper.

\section{COVERAGE METRIC BASED SOILING DETECTION TASK} \label{sec:soiling}

\subsection{Motivation}

As discussed in the previous section, soiling detection is a crucial algorithm for a safe autonomous driving system. A semantic segmentation task is a standard way to solve this problem. However, in this work, we aim to look for a more efficient solution. We use our extensive study on the design of efficient segmentation decoders \cite{dasvisapp19} and design a light decoder with 10X less runtime to suit embedded deployment. Besides, semantic segmentation requires excellent annotation of boundaries, and our proposed solution can handle coarse annotations.


Previous work on soiling detection ~\cite{uricar2019soilingnet, das2019soildnet} had tile-level classification. However, it was based on outputting the dominant soiling class (for example: clean, transparent, semi-transparent, opaque) that occupies most of the region in each tile. If the majority of the tiles are predicted as any class other than clean, then the cleaning system is activated to remove the soiled objects from the camera lens. Brief details on the cleaning system can be found in \cite{uricar2019soilingnet}.

However, the proposed detection system can fail in real-world scenarios. One such example is when the network has the respective confidence scores for a tile of e.g., clean-47\%, transparent-5\%, semitransparent-13\%, and opaque-35\%. In this case, it assumes that the cleaning system is not required as the clean class's high confidence indicates a mostly clean image.  This scenario shows that a simple classification scheme based on the maximum score is not ideal for evaluating the problem. Hence, there is a need for a more meaningful metric that represents the situation better.

\subsection{Class Coverage: Formal definition}

U\v{r}i\v{c}\'{a}\v{r} et. al~\cite{uricar2019soilingnet} formulated the soiling detection task as a multi-label holistic classification problem. However, we are using a tile-based classification, which helps to determine the degraded parts of the image and can be used to adopt the further processing pipeline accordingly. We provide the formal definition of the tile-based soiling classification task in the following paragraph. 

The input image $\mathcal{I}^{H \times W}$ of width $W$ and height $H$ is split into equally sized tiles, the number of tiles is parametric ($\mathrm{vtiles} \times \mathrm{htiles}$) and their size ($t_h$, $t_w$) is calculated as $t_h = \frac{H}{\mathrm{vtiles}}$, $t_w = \frac{W}{\mathrm{htiles}}$. Note, that we always set the number of both vertical and horizontal tiles in such a way that both $t_h, t_w \in \mathbb{N}$. The coverage is calculated based on the polygonal annotation of soiling, simply as the fraction of pixels from the given class $\{\mathrm{clean}, \mathrm{transparent}, \mathrm{semi-transparent}, \mathrm{opaque}\}$ that belong to the particular tile and the number of pixels of the tile itself. This definition ensures that the coverages of all classes within the tile sum up to $1$.

Finding the coverage of each class in a tile can be interpreted as a standard segmentation problem. However, for the application at hand, a precise segmentation mask is not necessary but only the area of the soiling segmentation. The percentage of the soiling classes in a tile is sufficient to decide on activating the cleaning system. So instead of solving the problem via a generic segmentation model, we explore to simplify it and use a smaller network motivated by embedded deployment. It can be analogous to using a more efficient bounding box detector instead of segmentation of objects. Bounding box detector indirectly captures a rough area of an object. However, explicitly finding areas of objects was not studied before, and it could be useful in other application domains like medical image processing. Hence we considered solving this problem using direct regression of area per tile. 


\section{DATASET DESIGN AND PROPOSED ARCHITECTURE} \label{sec:proposed}

\begin{figure}[t]
    \centering
    \includegraphics[width=0.5\textwidth]{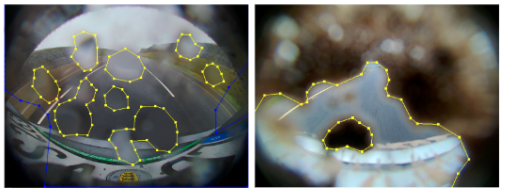}
    \includegraphics[width=0.49\textwidth]
    {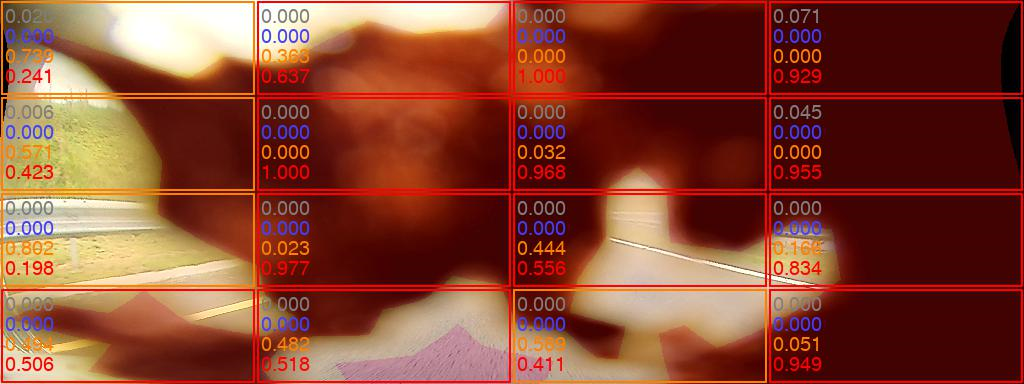}
    
    \caption{Soiling annotation using polygons (top) and tile level coverage values per class generation derived from polygons (bottom). }
    \label{fig:soilingannotation}
\end{figure}

The first subsection provides brief details on how the dataset is created and used in the experiment section. The later subsection explains the details of the CNN architecture (encoder and decoders) considered in this work.

\begin{figure*}
    \centering
    \includegraphics[width=0.75\textwidth]{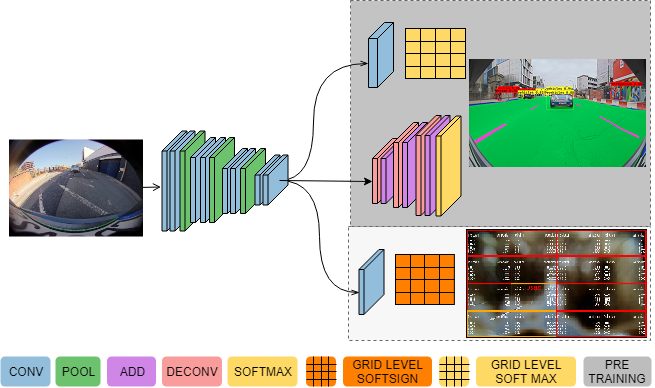}
    \caption{Illustration of soiling integrated into a multi-task object detection and segmentation network.}
    \label{fig:multi-stream-task}
\end{figure*}

\subsection{Dataset Creation}

Our dataset creation strategy remains the same as discussed in~\cite{uricar2019soilingnet, uricarvisapp19}. For this experiment, a total of $1,05,987$ images from all four cameras around the car was collected. We extracted every $15$-th frame out of short video recordings recorded at 30 frames per second. This sampling method was adapted to avoid highly similar frames in the dataset. 
The annotations were generated manually in the form of coarse polygonal segmentations. Finally, the polygon annotations were converted into tile level-based labels. The class label for each tile was determined based on the dominating soiling class coverage in the specific tile. Figure~\ref{fig:soilingannotation} shows an example of our polygon annotation (top) and its converted tile coverage annotation (bottom). The coverage values are mentioned in each tile in order clean (gray), transparent (blue), semitransparent (orange), and opaque (red).

The entire dataset was divided into three non-overlapping parts with $60$/$20$/$20$ ratio for training, validation, and test sets. A stratified sampling approach was followed to mostly retain the underlying distributions of the classes among the splits.

\subsection{Proposed CNN architecture}
The proposed network is a multitask learning framework with one shared encoder and dedicated decoders per task. This strategy has been established to be effective as it reduces the number of trainable parameters by a significant margin as it removes the need for task-specific encoders. For this experiment, we selected a ResNet-10 network without classification head as the encoder. Other than a soiling, there are two vision tasks - semantic segmentation and object detection. An FCN8~\cite{long2015fully} style decoder was adapted for semantic segmentation and the detection decoder is YOLO-V2~\cite{redmon2017yolo9000}. Under 2-task settings, our network architecture is similar to \cite{sistu2019real}. The soiling detection decoder is designed as a stack of convolution layers followed by batch normalization and ReLU activations. The overall architecture of our system is described in Figure~\ref{fig:multi-stream-task}. The network is implemented in the Keras framework \cite{chollet2015keras}. 

Our design goal is to integrate a soiling detection algorithm to an existing automated driving system with minimal computational overhead. We use a multitask strategy where we leverage existing encoder and build a light decoder motivated by \cite{sistu2019neurall} \cite{chennupati2019multinet++}. We first integrate soiling with the frozen encoder and only train soiling decoder following the auxiliary training strategy discussed in \cite{Chennupativisapp19}. We then use asynchronous backpropagation proposed in \cite{kokkinos2017ubernet}. 

First, detection and segmentation tasks are trained jointly. Then the encoder is frozen (set as non-trainable), and the soiling decoder is trained on top of it. The main reason for this is that our soiling dataset does not provide labels for the other tasks and that the soiling task itself is quite different than the other tasks. Therefore, we assume that joint training will likely disturb the more important detection and segmentation features during backpropagation.
The training was done batch-wise with a batch of size $64$ for $50$ epochs, and the initial learning rate was set to $0.001$ along with the Adam optimizer  \cite{kingma2014adam}. Categorical cross-entropy and categorical accuracy were used as loss and metrics respectively for the classification output.

This work's significant contribution includes the coverage metric that is very suitable for various real-world soiling scenarios. We propose an RMSE (Root Mean Square Error) value per class to measure the presence of each class per tile. The RMSE is defined as 

\begin{equation}
    \mathrm{RMSE} = \sqrt{\sum_{i\in[1,N]}\frac{1}{N}\sum_{j\in[1, C]}(t_{ij}-p_{ij})^2}\ ,
\end{equation}

\vspace{0.3cm}
in which $t$ and $p$ are the set of true and predicted values coming out of the softsign function. $\mathcal{C}$ is the set of classes $\mathcal{C} = \{ \mathrm{clean}, \mathrm{transparent}, \mathrm{semi-transparent}, \mathrm{opaque} \}$ and $N$ is the total number of tiles in one image. Here, $i$, $j$ $\in \mathcal{\mathbb{R}}$ and $i\in[1, N]$ and $j\in[1, C]$, respectively.  
The proposed coverage based RMSE and weighted precision were applied as loss and evaluation metrics for the coverage output.

\section{EXPERIMENTAL RESULTS} \label{sec:results}

\begin{figure*}[tb]
    \centering
    \includegraphics[width=0.3\linewidth]{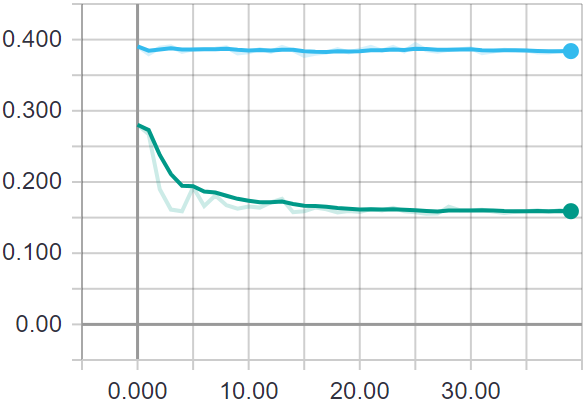}
    \includegraphics[width=0.3\linewidth]{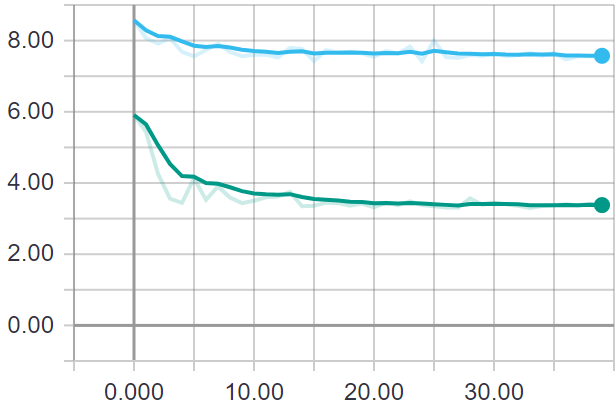}
    \includegraphics[width=0.3\linewidth]{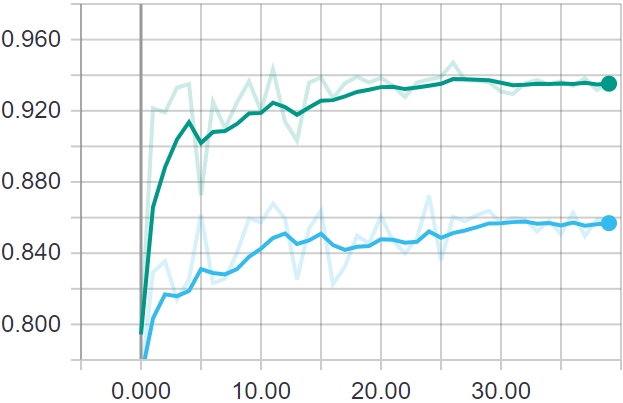}\\
    \hspace*{1cm}{$(a)$}\hspace*{5cm}{$(b)$}\hspace*{5cm}{$(c)$}\\
    \includegraphics[width=0.3\linewidth]{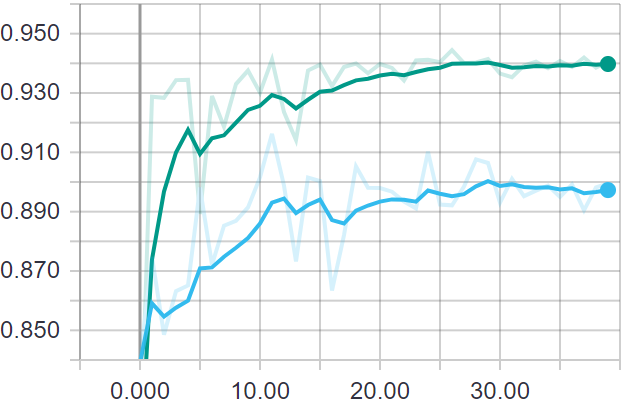}
    \includegraphics[width=0.3\linewidth]{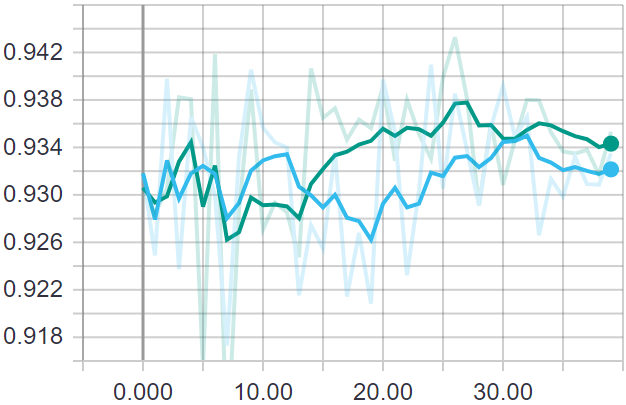}
    \includegraphics[width=0.3\linewidth]{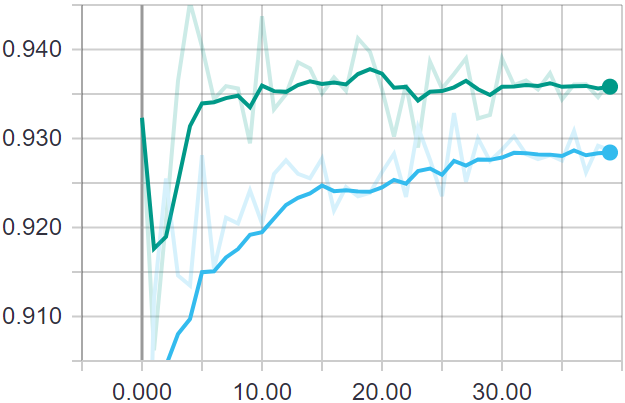}\\
    \hspace*{1cm}{$(d)$}\hspace*{5cm}{$(e)$}\hspace*{5cm}{$(f)$}\\
    \caption{Color codes: Blue/Green - trained with/without data augmentation and performance on validation dataset; From left to right: a) epochs vs. RMSE; b) epochs vs. weighted precision loss; c) epochs vs. accuracy for class clean; d) epochs vs. accuracy for class transparent; e) epochs vs. accuracy for class semitransparent; f) epochs vs. accuracy for class opaque}
    \label{fig:train_val_graph}
\end{figure*}

\begin{table*}[tb]
\centering
\resizebox{0.94\textwidth}{!}{
\begin{tabular}{|l|c|c|c|c|c|c|c|c|c|}
\hline


\multicolumn{2}{|c|}{Front camera} & \multicolumn{2}{c|}{Rear camera} & \multicolumn{2}{c|}{Left camera} & \multicolumn{2}{c|}{Right camera} & \multicolumn{2}{c|}{Overall} \\ \hline

{\textbf{Soiling Classes}} & \textbf{RMSE} & {\textbf{Soiling Classes}} & \textbf{RMSE} & {\textbf{Soiling Classes}} & \textbf{RMSE} & {\textbf{Soiling Classes}} & \textbf{RMSE} & {\textbf{Soiling Classes}} & \textbf{RMSE}\\ \hline
 Clean          & 0.1007    & Clean    & 0.1054    & Clean    & 0.1081    & Clean    & 0.1158    & Clean    & 0.1073\\ \hline
 Transparent    & 0.0941    & Transparent     & 0.0901    & Transparent     & 0.1151    & Transparent    & 0.0892    & Transparent    & 0.097\\ \hline
 Semitransparent  & 0.1194    & Semitransparent    & 0.1075    & Semitransparent    & 0.0842    & Semitransparent    & 0.0647    & Semitransparent    & 0.0948\\ \hline
 Opaque    & 0.1041    & Opaque    & 0.108    & Opaque    & 0.0966    & Opaque    & 0.0977    & Opaque    & 0.1017\\ \hline
\end{tabular}
}
\caption{Per class RMSE of tile level soiling detection (metrics rounded off) across camera views}
\label{tab:coverage}
\end{table*}

\begin{table*}[tb]
\centering
\resizebox{0.83\textwidth}{!}{
\begin{tabular}{|l||c|c|c|c||c|c|c|c|}
\hline
                       & \multicolumn{4}{c||}{Normalized Confusion Matrix}                                                                        & \multicolumn{4}{c|}{Raw Confusion Matrix}                                                                               \\ \hline
\multicolumn{9}{|c|}{\textbf{Training  on  all  cameras  and  Testing  on  all  cameras}}                                                                                                                                                                                                               \\ \hline
\multicolumn{1}{|c||}{--} & \multicolumn{1}{l|}{Clean} & \multicolumn{1}{l|}{Transparent} & \multicolumn{1}{l|}{Semitransparent} & \multicolumn{1}{l||}{Opaque} & \multicolumn{1}{l|}{Clean} & \multicolumn{1}{l|}{Transparent} & \multicolumn{1}{l|}{Semitransparent} & \multicolumn{1}{l|}{Opaque} \\ \hline
Clean                  & 0.91                          & 0.04                                & 0.02                           & 0.03                         & 113627                      & 5022                                & 2305                          & 3853                         \\ \hline
Transparent            & 0.01                       & 0.58                             & 0.24                           & 0.17                      & 169                        & 6543                             & 2704                          & 1895                        \\ \hline
Semitransparent                 & 0.01                       & 0.12                             & 0.70                        & 0.17                      & 117                        & 1302                              & 7476                        & 1810                       \\ \hline
Opaque                   & 0.01                          & 0.02                                & 0.09                        & 0.89                      & 226                          & 783                                & 3499                         & 35967                      \\ \hline
\multicolumn{9}{|c|}{\textbf{Training  on  all  cameras  and  Testing  on  front  cameras}}                                                                                                                                                                                                \\ \hline
\multicolumn{1}{|c||}{--} & \multicolumn{1}{l|}{Clean} & \multicolumn{1}{l|}{Transparent} & \multicolumn{1}{l|}{Semitransparent} & \multicolumn{1}{l||}{Opaque} & \multicolumn{1}{l|}{Clean} & \multicolumn{1}{l|}{Transparent} & \multicolumn{1}{l|}{Semitransparent} & \multicolumn{1}{l|}{Opaque} \\ \hline
Clean                  & 0.91 & 0.03 & 0.03 & 0.04                         & 31190 & 893 & 917 & 1213                \\ \hline
Transparent            & 0.00 & 0.54 & 0.35 & 0.10                      & 15 & 1619 & 1055 & 313                  \\ \hline
Semitransparent                 & 0.02 & 0.19 & 0.62 & 0.18                      & 88 & 811 & 2679 & 759               \\ \hline
Opaque                   & 0.00 & 0.01 & 0.08 & 0.90               & 47 & 85 & 832 & 9068                    \\ \hline
\multicolumn{9}{|c|}{\textbf{Training  on  all  cameras  and  Testing  on  rear  cameras}}                                                                                                                                                                                                \\ \hline
\multicolumn{1}{|c||}{--} & \multicolumn{1}{l|}{Clean} & \multicolumn{1}{l|}{Transparent} & \multicolumn{1}{l|}{Semitransparent} & \multicolumn{1}{l||}{Opaque} & \multicolumn{1}{l|}{Clean} & \multicolumn{1}{l|}{Transparent} & \multicolumn{1}{l|}{Semitransparent} & \multicolumn{1}{l|}{Opaque} \\ \hline
Clean                  & 0.93                          & 0.02                                & 0.03                           & 0.02                         & 27679 & 606 & 770 & 724              \\ \hline
Transparent            & 0.01 & 0.66 & 0.23 & 0.10            & 36 & 1925 & 685 & 284            \\ \hline
Semitransparent                 & 0.00 & 0.06 & 0.73 & 0.22             & 1 & 186 & 2457 & 737                 \\ \hline
Opaque                   & 0.02 & 0.00 & 0.14 & 0.84             & 138 & 41 & 1258 & 7673               \\ \hline

\multicolumn{9}{|c|}{\textbf{Training  on  all  cameras  and  Testing  on  left  cameras}}                                                                                                                                                                                                \\ \hline
\multicolumn{1}{|c||}{--} & \multicolumn{1}{l|}{Clean} & \multicolumn{1}{l|}{Transparent} & \multicolumn{1}{l|}{Semitransparent} & \multicolumn{1}{l||}{Opaque} & \multicolumn{1}{l|}{Clean} & \multicolumn{1}{l|}{Transparent} & \multicolumn{1}{l|}{Semitransparent} & \multicolumn{1}{l|}{Opaque} \\ \hline
Clean                  & 0.92                          & 0.05                                & 0.01                           & 0.02                         & 28356 & 1443 & 348 & 709              \\ \hline
Transparent            & 0.02 & 0.24 & 0.36 & 0.37            & 56 & 555 & 819 & 840            \\ \hline
Semitransparent                 & 0.01 & 0.07 & 0.84 & 0.08             & 9 & 69 & 853 & 83                 \\ \hline
Opaque                   & 0.00 & 0.05 & 0.07 & 0.88             & 35 & 567 & 742 & 9700               \\ \hline

\multicolumn{9}{|c|}{\textbf{Training  on  all  cameras  and  Testing  on  right  cameras}}                                                                                                                                                                                                \\ \hline
\multicolumn{1}{|c||}{--} & \multicolumn{1}{l|}{Clean} & \multicolumn{1}{l|}{Transparent} & \multicolumn{1}{l|}{Semitransparent} & \multicolumn{1}{l||}{Opaque} & \multicolumn{1}{l|}{Clean} & \multicolumn{1}{l|}{Transparent} & \multicolumn{1}{l|}{Semitransparent} & \multicolumn{1}{l|}{Opaque} \\ \hline
Clean                  & 0.88                          & 0.07                                & 0.01                           & 0.04                         & 26402 & 2080 & 270 & 1207              \\ \hline
Transparent            & 0.02 & 0.79 & 0.05 & 0.15            & 62 & 2444 & 145 & 458            \\ \hline
Semitransparent                 & 0.01 & 0.12 & 0.75 & 0.12             & 19 & 236 & 1487 & 231                 \\ \hline
Opaque                   & 0.00 & 0.01 & 0.06 & 0.93             & 6 & 90 & 617 & 9526               \\ \hline
\end{tabular}
}
\caption{Summary of results of tile-level soiling classification}
\label{tab:results}
\end{table*}

\begin{figure*}[tb]
    \centering
    \includegraphics[width=0.32\linewidth]{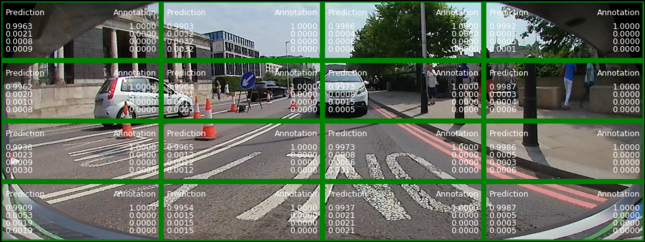}
    \includegraphics[width=0.32\linewidth]{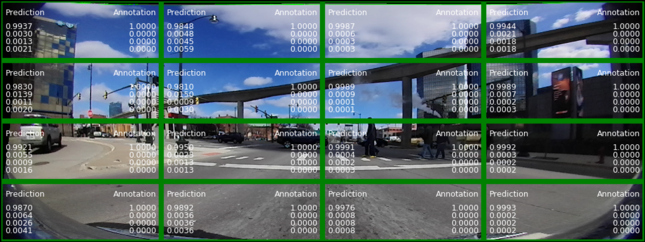}
    \includegraphics[width=0.32\linewidth]{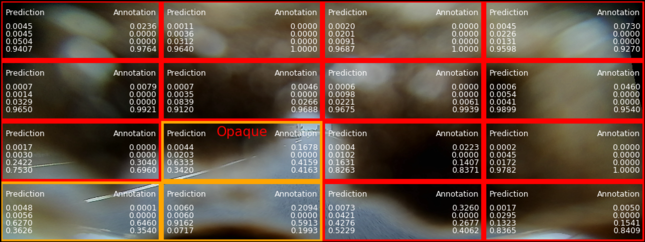}\\
    \vspace*{0.1cm}
    \includegraphics[width=0.32\linewidth]{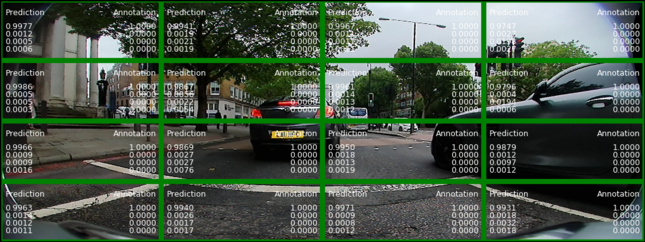}
    \includegraphics[width=0.32\linewidth]{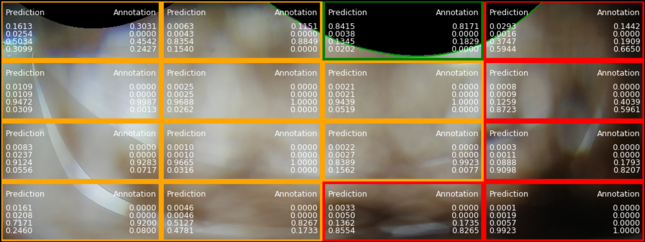}
    \includegraphics[width=0.32\linewidth]{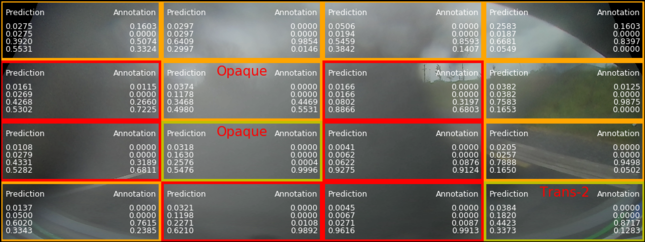}\\
    \caption{Results of tile based soiling detection. Misclassified tile(s) will have true label written as text in red. Color codes: Clean-Green, Transparent-Yellow, Semitransparent-Orange, Opaque-Red. Coverage values are best visible while in zoom.}
    \label{fig:sample_results}
\end{figure*}

In this section, we discuss and summarize our experimental results. Table~\ref{tab:coverage} presents the proposed coverage metric per class per camera view and overall. Table~\ref{tab:results} summarizes the tile level experiments on the four cameras individually and overall. Through Figure~\ref{fig:train_val_graph}, we discuss the adverse effect of data augmentation, particularly for this experiment.

Table~\ref{tab:coverage} presents the RMSE values per class for each camera view and overall. The achieved RMSE values for all camera views are fairly reasonable across all classes as the errors are bounded and quite close to $0$. These values are more meaningful than classification-based metrics because a tile will be represented by a coverage value for each class, which correlates with the occupancy of the same class within the tile. It further resolves the confusion between transparent and semitransparent classes highlighted in the next section. Figure~\ref{fig:sample_results} shows some examples of the proposed coverage results graphically where coverage predictions and annotations are noted per class in each tile.

\textbf{Tile Level Classification:} 
For this experiment, as mentioned earlier, tile-level soiling classification has been performed a number of tiles of $4 \times 4$ in an image. The camera view based confusion matrices for all soiling classes with and without data normalization are shown in Table \ref{tab:results}. From the table, it can be observed that the confusion is higher between transparent and semitransparent classes. It is likely the case due to a lack of variation in the dataset. It is only a subtle change in visibility that changes a tile from semitransparent to transparent. Furthermore, annotators are not always consistent across images as even humans struggle to annotate this kind of data.

\textbf{Data augmentation:}
Data augmentation is a base technique in vision-based applications. We applied some common augmentation techniques such as horizontal flips, contrast, brightness, and color (hue and saturation) modifications and adding Gaussian noise. These augmentations are controlled with probabilities to adapt them to the dataset. 
All samples were qualified for any of the augmentations based on the probability. However, it has been observed that augmentation is playing a negative role in this task. Figure~\ref{fig:train_val_graph} shows several loss and performance graphs captured during training on the validation dataset for two experiments of tile-based soiling detection - with and without augmentation. According to the trend across graphs, it can be seen that the test without augmentation outperformed the one with augmentation. We suppose a lack of scene variation in our dataset to explain these results as training without data augmentation favors by heart learning - which increases the score on the validation set. However, it can be expected that data augmentation will help a lot to generalize across different unseen weather and soiling conditions in practice.

\textbf{Practical Challenges:} 
The soiling detection may seem to look like a segmentation problem. The difficulty lies in defining the types of soiling at pixel level due to the unavailability of spatial or geometric structure in the soiling pattern. Moreover, some classes have less inter-class variance, such as transparent and semitransparent, which makes annotators' job extremely difficult. It is often a matter of discussion to judge whether specific pixels are transparent or semitransparent, especially in transition zones.

Furthermore, we noticed that cloud structures in the sky often confuse with the opaque soiling class. It commonly occurs in highway scenes due to large open areas of the sky being visible in the image containing unstructured, blurry patterns that seem very similar to soiling patterns. Sun glare \cite{sunglare} is another issue that makes some areas overexposed in the image leading to artifacts. These artifacts are often misclassified as opaque soiling.

\section{CONCLUSIONS} \label{sec:conc}
In this paper, we highlighted the current limitation of only classification based output for the soiling detection problem and how it impacts the system level in making decisions to clean the camera lens. As part of the solution, instead of pixel-level segmentation, we elaborated on an alternative and much reasonable metric that presents the problem in a more intuitive way. Furthermore, the tile-based approach is faster than semantic segmentation. The proposed coverage metric is useful to get control of each class within a tile. We experimented with our proposal in a Multi-Task Learning framework that contains semantic segmentation and object detection decoders where the training of the soiling decoder was performed following an asynchronous back-propagation algorithm. 
Future improvements will comprise the analysis of variation in the dataset to make data augmentation applicable to our system leading to better results. Furthermore, hyperparameter tuning (such as class weighting and regularization) as well as neural architecture search (NAS) will be considered to optimize both detection performance and run time on an embedded platform.
\bibliographystyle{ieeetr}
\bibliography{references/egbib}
\end{document}